\documentclass[journal]{IEEEtran}

\usepackage[table]{xcolor}
\usepackage{algorithm}
\usepackage{algorithmic}
\usepackage{amsmath,amssymb,amsfonts,mathrsfs}
\usepackage{array}
\usepackage{bm}
\usepackage{booktabs}
\usepackage{cite}
\usepackage{graphicx}
\usepackage{subfigure}
\usepackage{multirow}
\usepackage{threeparttable}
\usepackage{stfloats}
\usepackage{url}
\usepackage[colorlinks,linkcolor=blue,anchorcolor=blue,citecolor=blue]{hyperref}
\usepackage{cleveref}
\usepackage[T1]{fontenc}

\newcolumntype{L}[1]{>{\raggedright\arraybackslash}p{#1}}
\newcolumntype{C}[1]{>{\centering\arraybackslash}p{#1}}

\Crefname{table}{Table}{Tables}
\Crefname{equation}{Eq.}{Eqs.}
\Crefname{figure}{Fig.}{Figs.}
\Crefname{algorithm}{Algorithm}{Algorithms}
\Crefname{section}{Section}{Sections}
\Crefname{subsection}{Section}{Sections}
\graphicspath{{figures/}}

\newcommand{\ascendmoea}{AscendMOEA}
\newcommand{\code}[1]{\texttt{#1}}
\newcommand{\R}{\mathbb{R}}

\begin{document}
\raggedbottom

\title{AscendMOEA: An Ascend NPU-Accelerated Platform for Tensorized Multi-Objective Evolutionary Algorithms}

\author{
	Qianlong~Dang,~\IEEEmembership{Member,~IEEE,}
    Zhiyang~Zhang,
    Baosheng~Li,
	Tianqing~Zhu,~\IEEEmembership{Member,~IEEE,}
	Junhu~Ruan%
    \thanks{Qianlong~Dang, Zhiyang~Zhang, and Baosheng~Li are with the College of Science, Northwest A\&F University, Yangling 712100, China (e-mail: xidianqldang@163.com, jeanzhang@nwafu.edu.cn, baoshengli96@nwsuaf.edu.cn). 

Tianqing~Zhu is with the Faculty of Data Science, City University of Macau, Macau, China (e-mail: tqzhu@cityu.edu.mo). 

Junhu~Ruan is with the College of Economics and Management, Northwest A\&F University, Yangling 712100, China (e-mail: rjh@nwsuaf.edu.cn).}
}

%\markboth{IEEE Computational Intelligence Magazine}%
%{AscendMOEA}

\maketitle

\begin{abstract}
Multi-objective evolutionary optimization increasingly involves large populations, expensive problem evaluations, and diverse algorithmic structures. These workloads expose the limited parallelism of loop-based implementations, while rewrites for accelerator execution can alter selection order, tie handling, and population updates. AscendMOEA is presented as a scalable platform that executes multi-objective evolutionary algorithms (MOEAs) on Huawei Ascend neural processing units (NPUs) and conventional central processing units (CPUs). The platform represents the decision vectors, objectives, constraints, and auxiliary states as tensors and provides reusable interfaces for algorithms, problems, population operators, quality indicators, monitoring, and repeated experiments. Dense computations are executed in batches, such as dominance, distance, variation, and problem-evaluation calculations. Short recurrences with causal dependencies retain their prescribed order and consume precomputed tensor results, which preserves the behavior of the original algorithms. AscendMOEA contains 18 algorithms, 155 benchmark and application problems, 19 public operator interfaces, and 11 quality indicators. Experiments on eight Ascend NPUs and 192 physical CPU cores evaluate operator throughput, end-to-end runtime, memory consumption, numerical agreement, and implementation ablation. Tensorized operators obtain a median acceleration of 1.58 times on the NPU. At large population scales, complete algorithms achieve a median acceleration of 5.76 times while retaining the recorded quality indicators. Moreover, the results identify the workload scale and ordered computation that determine whether accelerator execution is beneficial. AscendMOEA is open-sourced at {\hypersetup{urlcolor=black}\url{https://github.com/dqlme/AscendMOEA}}.
\end{abstract}


\IEEEpeerreviewmaketitle

\section{Introduction}

\IEEEPARstart{M}{ulti-objective} optimization problems (MOPs) require a set of solutions that describes different compromises among conflicting objectives. Multi-objective evolutionary algorithms (MOEAs) provide a natural population-based approach for solving MOPs, which can approximate multiple regions of the Pareto front (PF) in one run~\cite{Zhou2011Survey}. This capability supports engineering design, resource allocation, scheduling, and data-driven applications. Recent studies address constrained multi-objective optimization problems (CMOPs), multimodal multi-objective optimization problems (MMOPs), sparse MOPs, and large-scale MOPs. These problem classes place different requirements on population representation and environmental selection. For example, MMOPs require multiple Pareto optimal sets in the decision space, and solving sparse MOPs needs to determine variable activation together with objective quality.

The expansion of algorithmic scope makes software infrastructure a substantive part of MOEA research. A reusable platform should coordinate algorithms, problems, operators, indicators, experiments, and visualization without embedding problem-specific assumptions into the execution process. Benchmarking repositories emphasize common instances and comparable settings. jMetal~\cite{Durillo2011JMetal}, ParadisEO-MOEO~\cite{Liefooghe2011ParadisEO}, and PlatEMO~\cite{Tian2017PlatEMO} establish broad component libraries in Java, C++, and MATLAB, respectively. Moreover, Python platforms provide direct access to scientific-computing tools. pymoo~\cite{Blank2020Pymoo} separates sampling, selection, variation, termination, and decision support into independent modules. jMetalPy~\cite{BenitezHidalgo2019JMetalPy} and pagmo~\cite{Biscani2020Pagmo} support parallel evaluation and heterogeneous solvers. These systems substantially improve implementation reuse. Nevertheless, population operators are still commonly expressed through individual-level loops or serial comparisons, and this limitation becomes prominent when objective evaluations are inexpensive.

Modern accelerators provide extensive parallel arithmetic, but an MOEA does not become accelerator efficient by moving an unchanged loop to another device. For example, pairwise dominance and distance calculations have abundant parallelism. In contrast, frequent scalar branches, small kernel launches, and device synchronization may consume more time than the associated arithmetic. The problem becomes more difficult when an update carries an intragenerational dependency. In MOEA/D~\cite{Zhang2007MOEAD}, a solution accepted for one subproblem can influence later subproblems in the same generation. Additionally, each deletion in distance-based truncation changes the surviving set considered by the next deletion in SPEA2~\cite{Zitzler2001SPEA2}. Simultaneously executing these decisions modifies the algorithm, even if the resulting code is faster.

Evolutionary systems for GPU execution address parts of this challenge. EvoJAX~\cite{Tang2022EvoJAX} batches neuroevolution workloads, and EvoX~\cite{Huang2025EvoX} provides a functional programming model with distributed GPU execution. EvoMO~\cite{Liang2026EvoMO} formulates representative MOEAs as tensor programs and demonstrates substantial acceleration at large scales. Its experimental discussion reports that batched niche selection and simultaneous MOEA/D updates can lead to trajectories that differ from the original algorithms. Consequently, acceleration and algorithmic equivalence need to be examined together. Similar terminal means alone cannot determine whether ranking, random events, function-evaluation counts, and ordered writes remain unchanged.

Another practical gap concerns accelerator diversity. Most current evolutionary libraries target central processing units (CPUs), NVIDIA graphics processing units (GPUs), or generic distributed clusters. Huawei Ascend NPUs employ the compute architecture for neural networks (CANN)~\cite{CANN2026} and the Da Vinci AI Core~\cite{Tang2023DaVinci}. Their matrix, vector, scalar, and data-movement units provide substantial throughput, while launch overhead, synchronization, operator coverage, and precision support differ from conventional GPU environments. Therefore, an MOEA platform for Ascend NPUs requires an explicit device policy rather than a direct replacement of the device string.

This study presents \ascendmoea{}, which is a tensorized and extensible platform for multi-objective evolutionary optimization on Ascend NPUs and conventional CPUs. The platform connects a batched problem interface with reusable population operators and tensor calculations defined for individual algorithms. Every run specifies its device, precision, random state, timing interval, monitor, and termination behavior. When an NPU is selected, all floating-point calculations adopt 32-bit floating-point precision (float32). Dense relations and population evaluations remain on the NPU. Short recurrences with causal dependencies operate on compact precomputed data and return only the resulting indices or masks. This organization reduces repeated synchronization and retains the update order of the original algorithms. Specifically, the main contributions of this study are summarized as follows:
\begin{enumerate}
\item An extensible software architecture is established for multi-objective evolutionary optimization on Ascend NPUs. A common population object connects 18 MOEAs and 155 benchmark or application problems with 19 public operator interfaces and 11 quality indicators. Typed interfaces and registries allow newly developed algorithms and problems to follow the same execution, monitoring, and experiment procedure as the built-in components.
\item A tensorization method is developed without changing the behavior prescribed by each algorithm. The method converts pairwise comparisons, reductions, selection, variation, and batched problem evaluation into tensor calculations. Validation requires exact agreement for selected indices, front numbers, masks, tie decisions, and function-evaluation counts. Numerical tolerances apply only to continuous values and are fixed according to the data type before the results are examined.
\item An execution strategy for Ascend NPUs coordinates dense tensor calculations with short recurrences that have causal dependencies. The NPU constructs expensive relations, and each ordered recurrence consumes only the compact tensors needed for its next decision. One workflow manages device initialization, precision, synchronization, random states, population histories, memory measurements, and terminal outcomes.
\item A matched evaluation protocol distinguishes the effects of common population operators, calculations inside each algorithm, and NPU execution. The experiments measure synchronized runtime and memory together with terminal indicators, convergence histories, and capacity outcomes. They cover unconstrained MOPs, CMOPs, MMOPs, sparse MOPs, large-scale MOPs, real-world MOPs, and combinatorial MOPs. The results identify the calculations that benefit from the NPU and the workloads that remain limited by sequential decisions.
\end{enumerate}

The remainder of this article is organized as follows. \Cref{sec:related} reviews software platforms and tensor execution for evolutionary computation. \Cref{sec:method} describes the design and implementation of \ascendmoea{}. \Cref{sec:experiments} presents the experimental design, analyzes the results, and discusses their practical implications and limitations. Finally, \Cref{sec:conclusion} summarizes the findings and outlines future research.

\section{Related Studies}\label{sec:related}

\subsection{Software Platforms for Multi-Objective Optimization}

The design of an optimization platform determines which parts of an experiment can be reused and compared. jMetal~\cite{Durillo2011JMetal} represents algorithms, problems, operators, and quality indicators through object-oriented abstractions. ParadisEO-MOEO~\cite{Liefooghe2011ParadisEO} constructs fine-grained components for fitness assignment, diversity preservation, and elitism. PlatEMO~\cite{Tian2017PlatEMO} integrates a broad collection of MOEAs, benchmark problems, indicators, and visual experiments within a unified MATLAB environment. Its algorithm and problem classes provide an influential interface for multi-objective evolutionary research. These platforms illustrate that a stable separation between algorithm logic and problem evaluation improves both research reuse and practitioner access.

Python software follows several complementary directions. pymoo~\cite{Blank2020Pymoo} provides a modular hierarchy for multi-objective optimization and decision support. jMetalPy~\cite{BenitezHidalgo2019JMetalPy} brings the jMetal design to Python and supports distributed evaluation. DEAP~\cite{Fortin2012DEAP} supplies lightweight evolutionary types and operators, while pagmo~\cite{Biscani2020Pagmo} organizes heterogeneous optimization algorithms through an island model. PyPop7~\cite{Duan2024PyPop7} focuses on population-based black-box optimization under a unified application programming interface. Most of these systems parallelize independent evaluations or complete runs. Fine-grained population relations remain a separate computational concern.

Reproducible experiments require stable configurations, run records, and statistical analysis. SimOpt~\cite{Eckman2023SimOpt} separates simulation models, optimization problems, and repeated experiments. PyPRS~\cite{Huang2026PyPRS} supports parallel ranking-and-selection procedures and preserves the information required to compare repeated runs. These capabilities complement component reuse because performance claims for an accelerator depend on matched workloads and synchronized measurements.

\subsection{Tensor Execution for Evolutionary Computation}

Parallel MOEAs are commonly organized through master--slave evaluation, islands, fine-grained populations, or hybrid structures~\cite{Talbi2019Parallel}. Parallel evaluation is effective when the objective function dominates runtime. When objectives are inexpensive, non-dominated sorting, distance construction, and environmental selection can retain a quadratic serial component. Aguilar-Rivera~\cite{AguilarRivera2020GPU} develops a vectorized implementation of NSGA-II~\cite{Deb2002NSGAII} with stochastic sorting and grid-based crowding. This study demonstrates that population relations contain exploitable parallelism, although modifications to ranking and diversity rules can change the search behavior.

GPU studies extend parallelism beyond objective evaluation. Moreno et al.~\cite{Moreno2018NDS} compare the execution time and energy consumption of non-dominated sorting on GPUs and CPUs, and show that the benefit depends on the population structure and problem scale. Together with the vectorized NSGA-II study, this evidence motivates measurements of individual population operators instead of conclusions based only on the requested device.

Tensor libraries provide a concise representation of batched population calculations. EvoJAX~\cite{Tang2022EvoJAX} applies JAX to hardware-accelerated neuroevolution. EvoX~\cite{Huang2025EvoX} extends this direction through functional states, workflow analysis, and distributed execution with JAX or Ray. EvoMO~\cite{Liang2026EvoMO} develops a general tensorization methodology and applies it to NSGA-III, MOEA/D, and HypE. Its results show that large populations can make accelerator execution highly effective. However, its tensorized MOEA/D evaluates subproblem updates as independent batches, and its tensorized NSGA-III employs batched random operations to approximate sequential niche selection. The resulting indicator values remain comparable in many experiments, but the computational process is no longer identical to the original implementation. \ascendmoea{} addresses a different objective. It retains the original update semantics and tensorizes the calculations that can be exchanged without changing the state transition.

PyTorch~\cite{Paszke2019PyTorch} provides the tensor interface of \ascendmoea{}, while TorchNPU~\cite{AscendPyTorch2026} maps supported operations to Ascend devices through CANN~\cite{CANN2026}. NumPy~\cite{Harris2020NumPy} and SciPy~\cite{Virtanen2020SciPy} support host-side data preparation and reference assets. Numba~\cite{Lam2015Numba} compiles short deterministic recurrences that have limited tensor parallelism. The Da Vinci AI Core combines matrix, vector, scalar, and data-movement units, and its effective throughput depends on arithmetic intensity and memory traffic~\cite{Tang2023DaVinci}. The law formulated by Amdahl~\cite{Amdahl1967} limits complete-run acceleration when ordered computation remains substantial. The Roofline model~\cite{Williams2009Roofline} further explains why low-intensity kernels can remain bounded by launch or memory costs. Accordingly, \ascendmoea{} evaluates the complete synchronized workflow rather than reporting kernel throughput alone.

\begin{table*}[t]
\centering
\footnotesize
\caption{Representative software for multi-objective optimization and accelerator execution. The final column summarizes the validation emphasized in the corresponding publication.}
\label{tab:platforms}
\resizebox{0.96\textwidth}{!}{%
\begin{tabular}{L{0.12\textwidth}L{0.17\textwidth}L{0.23\textwidth}L{0.22\textwidth}L{0.21\textwidth}}
\toprule
System & Primary scope & Software organization & Parallel or accelerator support & Reported validation \\
\midrule
PlatEMO~\cite{Tian2017PlatEMO} & Evolutionary multi-objective optimization & Algorithms, problems, indicators, experiments, and visualization in MATLAB & Parallel evaluation through the MATLAB ecosystem & Algorithm behavior and benchmark experiments \\\addlinespace[4pt]
pymoo~\cite{Blank2020Pymoo} & Multi-objective optimization in Python & Modular sampling, selection, variation, termination, and decision support & Parallel problem evaluation and vectorized numerical components & Optimization quality and software examples \\\addlinespace[4pt]
EvoX~\cite{Huang2025EvoX} & General scalable evolutionary computation & Functional states, workflows, analyzers, and execution engines & Distributed GPU execution through JAX and Ray & System scalability and model performance \\\addlinespace[4pt]
EvoMO~\cite{Liang2026EvoMO} & Tensorized evolutionary multi-objective optimization & Tensor formulations of representative algorithms & Large-population GPU execution & Runtime and indicator-level comparison \\\addlinespace[4pt]
\ascendmoea{} & Broad multi-objective evolutionary optimization & Algorithms, problems, operators, indicators, monitors, and typed registries & Ascend NPU and CPU execution with ordered host control & Operator equivalence, generation state, terminal indicators, runtime, and memory \\
\bottomrule
\end{tabular}}
\end{table*}

\section{Design and Implementation}\label{sec:method}

\subsection{Optimization Model and Platform Structure}

The architecture of \ascendmoea{} follows four objectives. Firstly, a problem should evaluate an entire population without knowing which algorithm requests the evaluation. Secondly, an algorithm should compose common operators without managing device initialization or measurement. Thirdly, the device policy should remain visible so that CPU and NPU experiments have an explicit precision and synchronization scope. Finally, a custom component should enter the same registry, workflow, monitoring, and experiment process as a built-in component. These objectives prevent accelerator details from being scattered throughout algorithm implementations.

A minimization MOP is formulated as follows:
\begin{equation}\label{eq:mop}
\begin{aligned}
 \min_{\bm{x}\in\Omega}\quad &\bm{f}(\bm{x})=
 \left(f_1(\bm{x}),\ldots,f_M(\bm{x})\right),\\
 \mathrm{s.t.}\quad &g_k(\bm{x})\leq 0,\qquad k=1,\ldots,K,
\end{aligned}
\end{equation}
where $\bm{x}\in\R^D$ is a decision vector, $D$ is the number of decision variables, $M$ is the number of objectives, and $K$ is the number of inequality constraints. The feasible decision region is denoted by $\Omega$. Constraint handling employs the total violation, which can be expressed as follows:
\begin{equation}\label{eq:violation}
 v(\bm{x})=\sum_{k=1}^{K}\max\{g_k(\bm{x}),0\}.
\end{equation}
A feasible solution is preferred to an infeasible solution, and two infeasible solutions are compared by $v$. For two feasible solutions $\bm{x}_a$ and $\bm{x}_b$, $\bm{x}_a$ Pareto dominates $\bm{x}_b$ when it is no worse in every objective and strictly better in at least one objective. A feasible solution that is not dominated by another feasible solution belongs to the Pareto optimal set, and its objective vector belongs to the PF.

\begin{figure*}[t]
\centering
\includegraphics[width=0.98\textwidth]{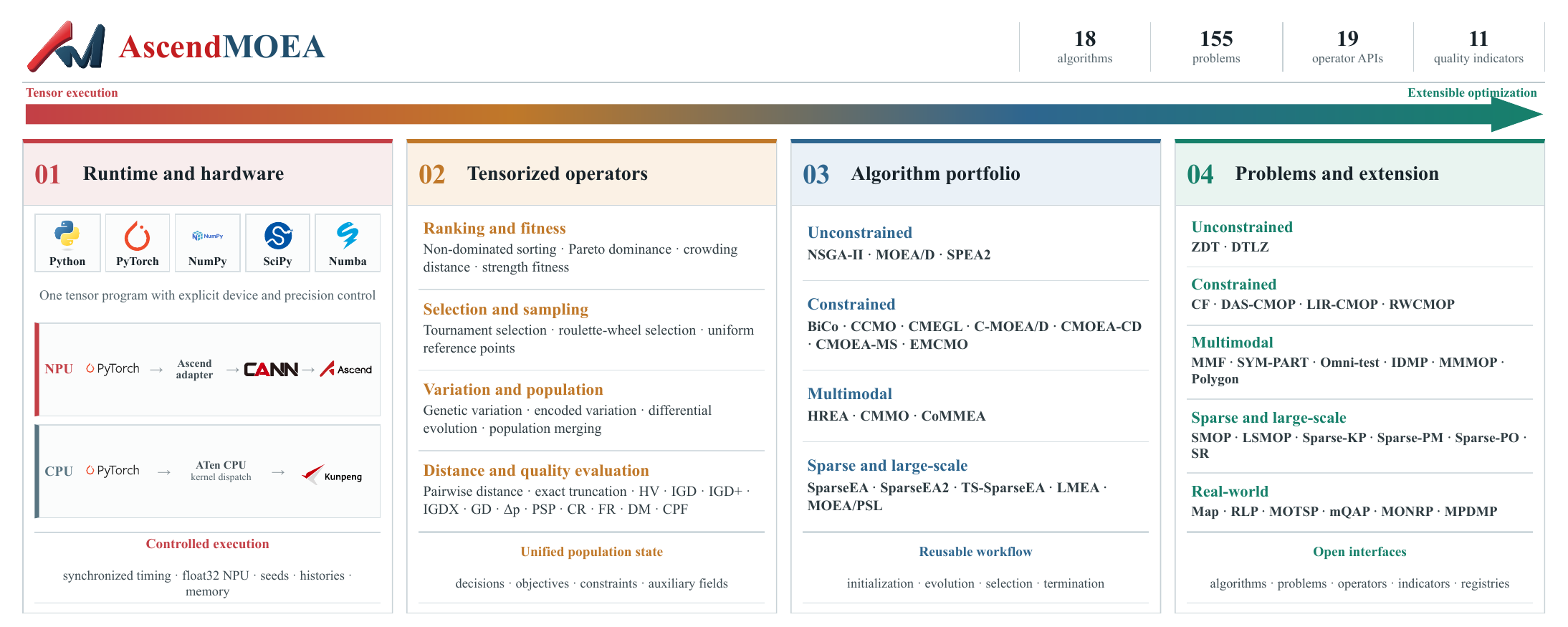}
\caption{Layered architecture and component portfolio of \ascendmoea{}.}
\label{fig:architecture}
\end{figure*}

For a population of size $N$, \ascendmoea{} stores the core state as follows:
\begin{equation}\label{eq:population}
\begin{aligned}
 \mathcal{P}&=(\bm{X},\bm{F},\bm{C}),\\
 \bm{X}&\in\R^{N\times D},\quad
 \bm{F}\in\R^{N\times M},\quad
 \bm{C}\in\R^{N\times K}.
\end{aligned}
\end{equation}
The rows of $\bm{X}$ contain decision vectors, and the corresponding rows of $\bm{F}$ and $\bm{C}$ contain objective and constraint values. Every tensor shares the same first dimension. This invariant allows a population to be selected, merged, cloned, detached, or transferred between devices through one object. \Cref{fig:architecture} presents the layered architecture of AscendMOEA. The runtime layer configures Python, PyTorch~\cite{Paszke2019PyTorch}, NumPy~\cite{Harris2020NumPy}, SciPy~\cite{Virtanen2020SciPy}, Numba~\cite{Lam2015Numba}, CANN~\cite{CANN2026}, the Ascend adapter~\cite{AscendPyTorch2026}, and Kunpeng CPU execution. The operator layer exposes ranking, fitness, selection, sampling, variation, population, distance, truncation, and indicator calculations. The algorithm layer combines these operators with state transitions defined by each MOEA. The upper layer contains numerical test suites, engineering applications, combinatorial instances, and extension interfaces. A single population representation connects the layers, which reduces conversion and avoids separate CPU and NPU implementations of the same MOEA.

The \code{Problem} class defines bounds, encodings, initialization, repair, objective evaluation, constraint evaluation, and reference data. Its public evaluator accepts a decision tensor with shape $(N,D)$. A numerical problem returns objective and constraint tensors as follows:
\begin{equation}\label{eq:batch_eval}
 (\bm{F},\bm{C})=\mathcal{E}(\bm{X}),\qquad
 \mathcal{E}:\R^{N\times D}\rightarrow
 \R^{N\times M}\times\R^{N\times K}.
\end{equation}
Real, integer, label, binary, and permutation encodings share this entry point. The generic evaluator enforces bounds and basic encoding validity before a problem-specific repair is applied. Objective and constraint values are then packed into a new population, and the number of function evaluations is updated by the batch size.

The \code{Algorithm} class retains the evolutionary \code{while} loop. Its termination method records the current population, invokes callbacks, and checks the function-evaluation budget. This arrangement preserves the control defined by each MOEA without duplicating lifecycle logic. The \code{Workflow} binds one MOEA and one MOP to a \code{RunConfig}. This configuration contains the device, random seed, CPU thread count, and synchronization policy used for timing. A \code{HistoryMonitor} can store every complete generation on the host, which prevents population histories from occupying accelerator memory. Additional monitors can process progress, diagnostics, or errors through the same lifecycle callbacks.

Typed component registries provide stable public names for algorithms and problems. A registry accepts a class directly or through a decorator, validates the component type, and creates instances from keyword arguments. The same mechanism supports third-party extensions. Consequently, a custom problem can be combined with any compatible algorithm, and a custom algorithm can directly import population operators. Both extensions inherit the standard seed, precision, monitoring, timing, and termination behavior.

\subsection{Tensorized Population Calculations}

Tensorization is applied only when the resulting state transition remains consistent with the original algorithm. Let $\mathcal{G}_{{r}}$ and $\mathcal{G}_{{t}}$ denote one generation of a source-consistent implementation and its tensorized implementation. Both calculations receive the same population $\mathcal{P}_t$, scalar state $s_t$, and explicit random-event tensor $\bm{\xi}_t$ as follows:
\begin{equation}\label{eq:generation_map}
 (\mathcal{P}_{t+1}^{q},s_{t+1}^{q})=
 \mathcal{G}_{q}(\mathcal{P}_t,s_t,\bm{\xi}_t),\qquad
 q\in\{{r},{t}\}.
\end{equation}
The function $\mathcal{I}$ extracts discrete outputs, including selected indices, front numbers, masks, archive membership, evaluation counts, and ordered writes. The function $\mathcal{Y}$ extracts floating-point outputs. Semantic preservation requires
\begin{equation}\label{eq:semantics}
 \mathcal{I}_{{t}}=\mathcal{I}_{{r}},\qquad
 \|\mathcal{Y}_{{t}}-\mathcal{Y}_{{r}}\|_{\infty}
 \leq \epsilon_{{a}}+
 \epsilon_{{r}}\|\mathcal{Y}_{{r}}\|_{\infty},
\end{equation}
where $\epsilon_{{a}}$ and $\epsilon_{{r}}$ are absolute and relative tolerances fixed for the data type before validation. A tolerance cannot justify a different tie decision, random event, feasibility state, or termination count. This distinction is essential because a small objective difference can still produce a different discrete population.

\begin{figure*}[t]
\centering
\includegraphics[width=0.98\textwidth]{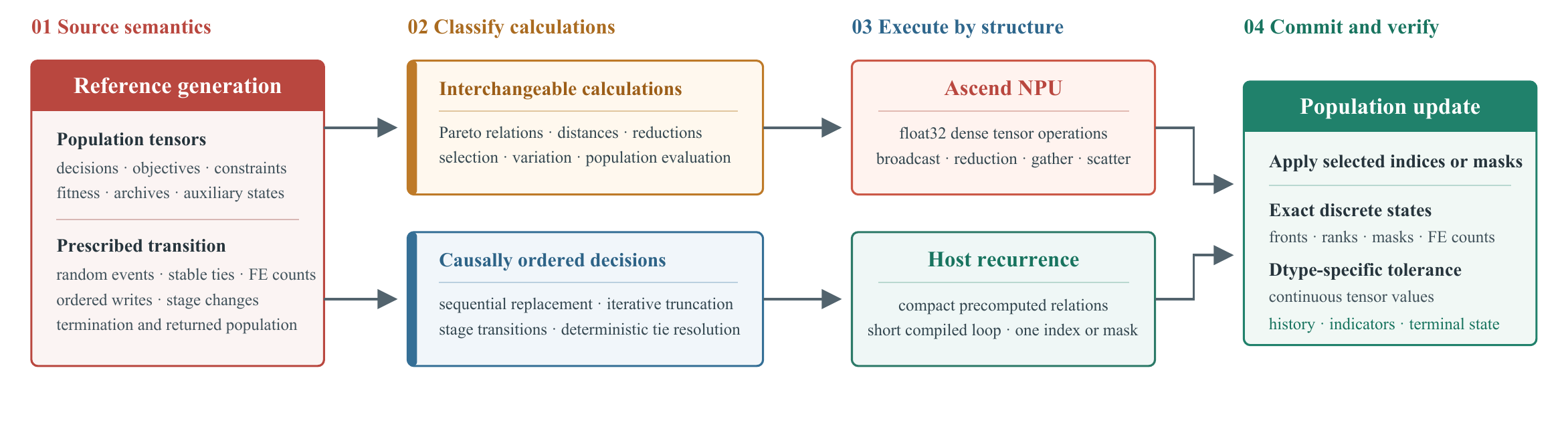}
\caption{Execution and validation process for semantics-preserving tensorization. Interchangeable population calculations are executed as dense NPU tensors, while calculations with causal dependencies retain their prescribed order.}
\label{fig:semantic_execution}
\end{figure*}

\Cref{fig:semantic_execution} shows how this criterion guides implementation. The original algorithm first defines the population state and the complete state transition. Then, calculations are separated according to whether their execution order can be exchanged. Pairwise relations, sorting reductions, selection, variation, and population-level problem evaluation enter the interchangeable group. Sequential replacements, iterative truncation, stage transitions, and deterministic tie resolution enter the ordered group. This classification follows the dependency between values rather than the apparent size of a loop.

The interchangeable group is evaluated through dense float32 tensor kernels when an Ascend NPU is selected. The ordered group receives compact tensors that have already been constructed by the device and executes only the recurrence required by the original algorithm. Both groups contribute to one population update. The updated indices, fronts, masks, evaluation count, and history are then compared with the source transition. This process keeps validation close to the generation boundary and makes the reason for each host-controlled calculation explicit.

Many population operators compare every pair of individuals. Under minimization, the objective value of individual $i$ on objective $m$ is denoted by $F_{im}$. Reshaping $\bm{F}$ to $(N,1,M)$ and $(1,N,M)$ allows broadcasting to construct the Pareto dominance relation employed by NSGA-II~\cite{Deb2002NSGAII}, which can be expressed as follows:
\begin{equation}\label{eq:dominance}
 D_{ij}=
 \left[\bigwedge_{m=1}^{M}F_{im}\leq F_{jm}\right]
 \wedge
 \left[\bigvee_{m=1}^{M}F_{im}<F_{jm}\right].
\end{equation}
The binary value $D_{ij}=1$ indicates that individual $i$ dominates individual $j$. Summation along the first dimension gives the domination count $c_j=\sum_iD_{ij}$. The first non-dominated front contains the indices for which $c_j=0$. After one front is labeled, the corresponding row contributions are subtracted from the remaining domination counts. The necessary loop proceeds over fronts rather than individuals, while comparisons, reductions, and mask updates remain tensor operations.

Constrained sorting first identifies infeasible individuals and calculates the sum of positive constraint violations. Their objective vectors are shifted beyond the current objective maximum according to the violation value. Then, the same dominance calculation can reproduce the feasibility rule applied by the original implementation~\cite{Deb2002NSGAII}. This formulation retains a single ranking operator for unconstrained and constrained populations.

Within a front $\mathcal{F}$, stable sorting produces the objective order required by crowding-distance calculation~\cite{Deb2002NSGAII}:
$f_{(1)m}\leq\cdots\leq f_{(|\mathcal{F}|)m}$ for objective $m$. Boundary individuals receive an infinite crowding distance. The contribution of an interior individual is defined as follows:
\begin{equation}\label{eq:crowding}
 q_{(i)m}=
 \frac{f_{(i+1)m}-f_{(i-1)m}}
 {f_{(|\mathcal{F}|)m}-f_{(1)m}},\qquad
 i=2,\ldots,|\mathcal{F}|-1.
\end{equation}
All objective columns are sorted together. Adjacent differences are calculated in the sorted order, and a scatter operation returns each contribution to the original individual index. Stable sorting preserves deterministic ties. Zero-range and non-finite cases follow the original behavior instead of being silently replaced by an alternative crowding rule.

\Cref{alg:ndselection} summarizes how these relations support environmental selection. The loop advances by complete fronts, while every pairwise comparison and every update within one front remains a tensor calculation. Only the final front requires crowding-based truncation.

\begin{algorithm}[t]
\footnotesize
\caption{Tensorized Non-Dominated Selection}
\label{alg:ndselection}
\begin{algorithmic}[1]
\REQUIRE $\bm{F}\in\R^{N\times M}$ and survivor count $n_{\mathrm{keep}}$.
\ENSURE Survivor index tensor $\bm{s}$.
\STATE Broadcast $\bm{F}$ and calculate the dominance matrix $\bm{D}$ by \Cref{eq:dominance}.
\STATE Calculate domination counts $\bm{c}\gets\bm{1}^{\mathrm{T}}\bm{D}$.
\STATE Initialize $\bm{s}$ as empty and $\bm{r}\gets\bm{c}$.
\WHILE{$|\bm{s}|<n_{\mathrm{keep}}$}
  \STATE Form the next front $\bm{h}\gets\{j\mid r_j=0\}$ by a Boolean mask.
  \IF{$|\bm{s}|+|\bm{h}|\leq n_{\mathrm{keep}}$}
    \STATE Append $\bm{h}$ and update $\bm{r}\gets\bm{r}-\bm{1}^{\mathrm{T}}\bm{D}_{\bm{h},:}$.
    \STATE Mark the appended entries so that they cannot enter another front.
  \ELSE
    \STATE Calculate crowding distances within $\bm{h}$ by stable sorting and scatter.
    \STATE Append the required entries of $\bm{h}$ in descending crowding order.
  \ENDIF
\ENDWHILE
\STATE Return $\bm{s}$ in its deterministic tie order.
\end{algorithmic}
\end{algorithm}

Tournament selection~\cite{Deb2002NSGAII} stores $T$ candidates for every required parent in an index tensor with shape $(T,N)$. Fitness values are gathered for all tournaments, and stable sorting applies the criteria in lexicographic order. Roulette selection normalizes inverse fitness forms a cumulative sum and applies \code{searchsorted} to a batch of uniform random values. Reference points are generated from simplex lattices and cached by population size, objective count, and data type.

Real-valued genetic variation processes all parent pairs simultaneously. Simulated binary crossover and polynomial mutation construct random, distribution, boundary, and activation tensors with shape $(N,D)$. Differential evolution batches base vectors, difference vectors, scale factors, and crossover masks~\cite{Storn1997DE}. Encoded variation applies column masks for integer, label, binary, permutation, and mixed decision representations. The associated repair is evaluated by batch. These calculations preserve the random-event layout so that controlled operator tests can compare offspring entry by entry.

Pairwise Euclidean distance uses direct broadcasting for moderate inputs. For larger matrices $\bm{X}$ and $\bm{Y}$, squared distance is calculated as follows:
\begin{equation}\label{eq:distance}
 \|\bm{x}_i-\bm{y}_j\|_2^2=
 \|\bm{x}_i\|_2^2+\|\bm{y}_j\|_2^2-
 2\bm{x}_i^{\mathrm{T}}\bm{y}_j.
\end{equation}
The inputs are centered before matrix multiplication. Entries that approach the roundoff threshold are recomputed through direct element-wise differences, and small negative values are clamped to zero. This correction avoids the cancellation that changes nearest-neighbor order for nearly identical points.

Problem evaluation follows the same batching principle. Separable terms are calculated for all individuals before reduction, while constraint values form a matrix with the population as its first dimension. Multimodal test problems calculate distances from all individuals to multiple reference centers through a three-dimensional tensor. Sparse problems reuse variable masks and group indices across the population. Permutation and assignment problems apply batched gather operations over route or assignment positions. As a result, objective evaluation no longer reintroduces an individual-level Python loop after the population operators have been tensorized.

\Cref{alg:offspring} gives the common offspring procedure. Random events are generated before variation and retain the same tensor positions in reference and tensorized tests. Repair and problem evaluation then process the complete offspring population.

\begin{algorithm}[t]
\footnotesize
\caption{Batched Variation and Problem Evaluation}
\label{alg:offspring}
\begin{algorithmic}[1]
\REQUIRE Parent population $\mathcal{P}$, parent indices $\bm{p}$, random tensor $\bm{\xi}$, and $Problem$.
\ENSURE Evaluated offspring population $\mathcal{P}_{\mathrm{o}}$.
\STATE Gather all parent tensors with $\bm{p}$.
\STATE Construct crossover, mutation, and activation masks from $\bm{\xi}$.
\STATE Calculate every offspring decision vector in one batched variation call.
\STATE Apply encoding-specific bounds and repair to the offspring tensor $\bm{X}_{\mathrm{o}}$.
\STATE Evaluate $(\bm{F}_{\mathrm{o}},\bm{C}_{\mathrm{o}})\gets\mathcal{E}(\bm{X}_{\mathrm{o}})$ by \Cref{eq:batch_eval}.
\STATE Form $\mathcal{P}_{\mathrm{o}}\gets(\bm{X}_{\mathrm{o}},\bm{F}_{\mathrm{o}},\bm{C}_{\mathrm{o}})$.
\STATE Increase $FE$ by the number of rows in $\bm{X}_{\mathrm{o}}$ and return $\mathcal{P}_{\mathrm{o}}$.
\end{algorithmic}
\end{algorithm}

\subsection{Ordered Updates and Device Execution}

Device initialization precedes the transfer of mutable problem states. CPU processes disable automatic loading of the accelerator backend before PyTorch~\cite{Paszke2019PyTorch} is imported, which prevents a CPU run from creating an unrelated NPU context. The NPU runtime is loaded only when the selected device requests it, and the active device is fixed before initialization, reference points, and problem data are generated.

The default real type is 64-bit floating-point precision (float64) on a CPU and float32 on an NPU. A matched CPU experiment can request float32. An NPU request for float64 is converted to float32 because several required Ascend kernels do not support float64. Floating-point calculations performed by host-controlled recurrences in an NPU run follow the same precision. Random seeds are propagated to Python, NumPy, PyTorch, and the active accelerator backend. Integer indices and Boolean masks retain their discrete types.

Timing follows the asynchronous execution of the NPU. The workflow synchronizes the selected device immediately before the timer starts and again after the algorithm finishes. The resulting interval includes every operation submitted by the optimization run and excludes unfinished work from an earlier calculation. Indicator calculation and result serialization are measured separately. Built-in and user-defined components share this timing interface.

Some calculations cannot be exchanged without changing the algorithmic process. MOEA/D~\cite{Zhang2007MOEAD} and C-MOEA/D~\cite{Jain2014CMOEAD} update subproblems sequentially. Once an offspring replaces one or more neighbors, a later subproblem observes the updated population. A simultaneous write would cause every subproblem to read the same population at the beginning of the generation. Therefore, \ascendmoea{} tensorizes weight generation, neighborhood construction, parent indexing, scalarizing values, and objective evaluation, while preserving the prescribed subproblem order.

SPEA2~\cite{Zitzler2001SPEA2} contains another ordered recurrence. The first deletion in distance truncation changes the neighbor sequence used to choose the second deletion. \ascendmoea{} calculates and stably sorts the complete distance relations on the tensor device. For an NPU run, the sorted distances and neighbor indices are transferred once to the host. Then, a Numba-compiled recurrence performs the deterministic deletions, and only the final retention mask returns to the NPU. This structure avoids one device synchronization per deletion and retains the exact truncation order.

\Cref{alg:generation} summarizes the resulting generation process. Dense population calculations stay on the selected tensor device. Ordered control receives precomputed relations and produces compact discrete outputs. The generational \code{while} loop remains visible in every algorithm, which keeps termination, parameter schedules, and state transitions close to their mathematical definitions.

\begin{algorithm}[t]
\footnotesize
\caption{Semantics-Preserving Tensorized Generation}
\label{alg:generation}
\begin{algorithmic}[1]
\REQUIRE $\mathcal{P}$ (population), $Problem$, $Device$, and $FE_{\max}$.
\ENSURE $\mathcal{P}_{out}$ (final population).
\STATE Initialize batched indices, neighborhoods, and random tensors on $Device$.
\WHILE{$FE<FE_{\max}$}
  \STATE Select parents through batched gather and stable tie rules.
  \STATE Generate, repair, and evaluate the complete offspring batch.
  \STATE Calculate interchangeable relations, distances, and fitness on $Device$.
  \IF{the environmental update contains causal writes}
    \STATE Execute ordered control with the precomputed tensors.
  \ELSE
    \STATE Update the population through batched masks and scatter.
  \ENDIF
  \STATE Record the complete generation and update $FE$.
\ENDWHILE
\STATE Synchronize $Device$ before returning the measured runtime.
\end{algorithmic}
\end{algorithm}

The synchronized runtime on device $d$ is decomposed as follows:
\begin{equation}\label{eq:runtime}
 T_d(N)=T^{\mathrm{fixed}}_d+
 T^{\mathrm{dense}}_d(N)+T^{\mathrm{ordered}}(N)+
 T^{\mathrm{move}}_d(N),
\end{equation}
where $d\in\{\mathrm{CPU},\mathrm{NPU}\}$. The term $T^{\mathrm{fixed}}_d$ contains launches, synchronization, and shape-specific compilation that occurs after timing begins. The dense tensor operations are represented by $T^{\mathrm{dense}}_d(N)$. The term $T^{\mathrm{ordered}}(N)$ measures causally ordered host control, and $T^{\mathrm{move}}_d(N)$ contains transfers between the tensor device and that control calculation. NPU execution becomes faster when
\begin{equation}\label{eq:crossover}
 T^{\mathrm{dense}}_{\mathrm{CPU}}(N)-
 T^{\mathrm{dense}}_{\mathrm{NPU}}(N)>
 \Delta T^{\mathrm{fixed}}+\Delta T^{\mathrm{move}},
\end{equation}
where each $\Delta$ is the NPU cost minus the corresponding CPU cost. The ordered term is shared when both devices execute the same recurrence. Hence, population size alone does not determine acceleration. The density of ranking, distance, variation, and problem evaluation needs to be considered together with the ordered fraction.

\subsection{Algorithms, Problems, and Extension}

The algorithm library covers four complementary categories, as summarized in \Cref{tab:portfolio}. The general MOEAs include NSGA-II~\cite{Deb2002NSGAII}, MOEA/D~\cite{Zhang2007MOEAD}, and SPEA2~\cite{Zitzler2001SPEA2}, which represent dominance, decomposition, and strength mechanisms. The constrained multi-objective evolutionary algorithms (CMOEAs) cover bidirectional coevolution, auxiliary tasks, staged constraint handling, and decomposition under constraints. The multimodal multi-objective evolutionary algorithms (MMOEAs) maintain diversity in the decision space and the objective space. Sparse and large-scale MOEAs manipulate variable masks, decision groups, or learned structures with fewer dimensions.

\begin{table*}[t]
\centering
\footnotesize
\caption{Algorithm and problem portfolio of \ascendmoea{}.}
\label{tab:portfolio}
\resizebox{\textwidth}{!}{%
\begin{tabular}{lcL{0.4\textwidth}L{0.4\textwidth}}
\toprule
Category & Count & MOEAs or MOP suites & Principal tensorized calculations \\
\midrule
General MOEAs & 3 & NSGA-II~\cite{Deb2002NSGAII}, MOEA/D~\cite{Zhang2007MOEAD}, and SPEA2~\cite{Zitzler2001SPEA2} & Dominance, fronts, crowding, scalarization, neighborhoods, strength fitness, and exact truncation \\\addlinespace[4pt]
CMOEAs & 7 & BiCo~\cite{Liu2022BiCo}, C-MOEA/D~\cite{Jain2014CMOEAD}, CCMO~\cite{Tian2021CCMO}, CMEGL~\cite{Qiao2023CMEGL}, CMOEA-CD~\cite{Liu2025CMOEACD}, CMOEA-MS~\cite{Tian2022CMOEAMS}, and EMCMO~\cite{Qiao2022EMCMO} & Constraint violation, feasibility masks, auxiliary populations, adaptive stages, and population exchange \\\addlinespace[4pt]
MMOEAs & 3 & HREA~\cite{Li2023HREA}, CMMO~\cite{Ming2023CMMO}, and CoMMEA~\cite{Li2023CoMMEA} & Local convergence, neighborhoods in the decision space, hierarchical ranking, epsilon relations, and archive cooperation \\\addlinespace[4pt]
Sparse and large-scale MOEAs & 5 & SparseEA~\cite{Tian2020SparseEA}, SparseEA2~\cite{Zhang2023SparseEA2}, TS-SparseEA~\cite{Jiang2022TSSparseEA}, LMEA~\cite{Zhang2018LMEA}, and MOEA/PSL~\cite{Tian2021MOEAPSL} & Variable scoring, sparse masks, grouped variation, correlation, and model training \\
\midrule
Unconstrained MOPs & 13 & ZDT1--6~\cite{Zitzler2000ZDT} and DTLZ1--7~\cite{Deb2002DTLZ} & Separable, disconnected, biased, degenerate, and many-objective PFs \\\addlinespace[4pt]
CMOPs & 33 & CF1--10~\cite{Zhang2008CF}, DASCMOP1--9~\cite{Fan2020DASCMOP}, and LIRCMOP1--14~\cite{Fan2019LIRCMOP} & Adjustable feasibility, disconnected regions, and large infeasible barriers \\\addlinespace[4pt]
MMOPs & 74 & MMF and CEC 2019~\cite{Liang2019CECMMOP}, IDMP~\cite{Liu2020IDMP}, MMMOP~\cite{Yue2019MMMOP}, Polygon~\cite{Liu2018Polygon}, and four multimodal applications & Equivalent and local Pareto sets, imbalanced basins, scattered optima, and geometric regions \\\addlinespace[4pt]
Sparse and large-scale MOPs & 17 & SMOP1--8~\cite{Tian2020SparseEA} and LSMOP1--9~\cite{Cheng2017LSMOP} & Sparse decision structures, variable linkage, and high decision dimensions \\\addlinespace[4pt]
Real-world and combinatorial MOPs & 18 & RWMOP1--10~\cite{Kumar2021RWCMOP}, sparse applications, MOTSP~\cite{Corne2007MOTSP}, mQAP~\cite{Knowles2003MQAP}, MONRP~\cite{Zhang2007MONRP}, and MPDMP~\cite{Koppen2007MPDMP} & Engineering constraints, routes, assignments, release planning, and discrete decisions \\
\bottomrule
\end{tabular}}
\end{table*}

Each MOEA remains in an independent Python module and inherits the same base class. Common operators are imported from the public operator package. Fitness, archive, and stage calculations that belong to one MOEA remain in its module, which keeps the implementation traceable to the source publication. The supplementary material provides the complete mapping between publications and tensorized calculations.

The 155 registered MOPs form five disjoint categories in \Cref{tab:portfolio}. Four multimodal applications are included with the MMOPs. The map-based problem~\cite{Ishibuchi2011Map} examines diversity maintenance through disconnected regions in the decision space. Rental location selection is introduced with a dual-population coevolutionary algorithm~\cite{Li2024RentalLocation}. RLP originates from the application study of bidirectional coevolution~\cite{Li2025MMEABDC}, and UECLP contains shifted local PFs~\cite{Zhang2025UECLP}. The category of real-world and combinatorial MOPs adds engineering design, sparse applications, routing, assignment, release planning, and decision problems with many objectives.

Reference PFs and Pareto sets are cached after construction. Numerical problems calculate objectives and constraints in batches, and fixed application data are stored with the package. Therefore, the same problem class executes on a CPU or NPU without maintaining separate formulas. This property is particularly important for a platform study. Otherwise, an accelerated algorithm can spend most of its runtime in a loop-based objective function.

A custom problem inherits \code{Problem}, defines dimensions and bounds in \code{setting}, and implements a batched \code{cal\_obj}. Optional methods provide constraints, repair, reference points, or a drawable PF. A custom algorithm inherits \code{Algorithm}, implements the evolutionary loop, and calls \code{not\_terminated} at generation boundaries. Decorators register either class under a stable public name. For example, a two-objective sphere problem can calculate both objectives by column-wise reductions over one decision tensor and can then be executed by \code{optimize(NSGAII(), problem, device="npu:0", seed=7)}. The same workflow returns the final population, evaluation count, device, and synchronized runtime. Detailed extension examples and public operator signatures are given in the supplementary material.

\section{Experiments and Analysis}\label{sec:experiments}

\subsection{Experimental Design and Validation}

The experiments address four research questions. RQ1 examines whether tensorized population operators preserve their reference outputs and identifies the scale at which NPU execution becomes advantageous. RQ2 evaluates the runtime and memory behavior of complete MOEAs across MOP categories and population scales. RQ3 determines whether CPU and NPU runs retain terminal indicators and convergence throughout evolution. RQ4 separates the contribution of replacing common operators from the contribution of tensorizing calculations inside each MOEA and evaluating each MOP in batches.

Three matched implementations are compared. The reference implementation retains source-consistent scalar or loop-based population calculations. The operator implementation replaces the common population operators but leaves calculations inside each MOEA unchanged. The full implementation additionally tensorizes these remaining calculations and evaluates each MOP in population batches. MOEA, MOP, seed, device, precision, and termination settings remain matched. Therefore, timing ratios can be associated with the corresponding implementation change.

Experiments run on a server with four Kunpeng 920 CPUs, 192 physical CPU cores, 1~TB host memory, and eight Ascend 910B4 NPUs. Each CPU task employs eight PyTorch~\cite{Paszke2019PyTorch} intra-operation threads. Independent CPU runs are scheduled as separate processes, while one persistent formal worker is bound to each NPU. Both CPU and NPU timing comparisons use float32. CPU processes disable accelerator-backend autoloading before PyTorch is imported. NPU timing synchronizes the device immediately before and after every measured interval.

The compute interval includes the complete optimization run and excludes indicator calculation, history serialization, and file writing. Operator configurations perform an unmeasured warmup before five timed repetitions. Standard algorithm experiments contain 20 paired seeds. Scale and ablation configurations contain five paired seeds when complete. Each run records process-tree resident set size (RSS), proportional set size for concurrent CPU processes, NPU allocated and reserved memory, and process-level high-bandwidth memory. Out-of-memory (OOM), timeout, device error, invalid indicator, and interruption are stored as distinct terminal states. A run that exceeds 14~h terminates with its elapsed time, completed evaluations, latest generation, and memory samples.

\Cref{tab:protocol} summarizes the evaluated workloads. Operator stress tests vary population size, objective count, decision dimension, duplicate rate, encoding, and selection configuration. The standard algorithm experiment uses $N=100$ and $FE_{\max}=10{,}000$. The scale experiment increases $N$ from $1{,}000$ to $20{,}000$ and increases $FE_{\max}$ proportionally from $1\times 10^5$ to $2\times10^6$. General MOEAs use ZDT or DTLZ problems in the scale tests. CMOEAs, MMOEAs, and sparse or large-scale MOEAs use MOPs from their corresponding suites.

\begin{table*}[t]
\centering
\footnotesize
\caption{Experimental workloads and recorded evidence.}
\label{tab:protocol}
\resizebox{0.99\textwidth}{!}{%
\begin{tabular}{L{0.18\textwidth}L{0.30\textwidth}L{0.17\textwidth}L{0.26\textwidth}}
\toprule
Experiment & Configurations & Repetitions & Recorded outputs \\
\midrule
Operator stress & $N\in\{1000,2000,5000,10000,20000\}$ with varied $M$, $D$, ties, encodings, and operator parameters & 5 after warmup & Runtime, throughput, host memory, NPU memory, output agreement, OOM, and timeout \\
Standard algorithms & $N=100$ and $FE_{\max}=10{,}000$ for every compatible algorithm--problem pair & 20 paired seeds & Runtime, complete population histories, indicators, FE, terminal state, and memory \\
Scale algorithms & $(N,FE_{\max})$ from $(1000,1\times10^5)$ to $(20000,2\times10^6)$ & 5 paired seeds & CPU-to-NPU speedup, convergence, memory growth, OOM, and timeout \\
Implementation ablation & Reference, common-operator, and full tensorized implementations under matched settings & 5 paired seeds & Reference/operator, operator/full, and reference/full timing gains \\
\bottomrule
\end{tabular}}
\end{table*}

Hypervolume (HV) measures the dominated objective-space volume~\cite{Zitzler1998HV}. Inverted generational distance (IGD) and modified inverted generational distance (IGD+) evaluate convergence and coverage, while inverted generational distance in the decision space (IGDX) evaluates approximation of the Pareto set~\cite{Ishibuchi2015IGDPlus,Li2019QualitySurvey}. The reciprocal of Pareto sets proximity, denoted by $1/\mathrm{PSP}$, combines decision-space coverage and IGDX with a minimization orientation~\cite{Yue2018PSP}. Constrained experiments additionally record feasible rate and constraint violation. Sparse experiments record the nonzero-variable ratio. Runtime and memory are summarized by medians, interquartile ranges, means, standard deviations, and 95\% bootstrap confidence intervals (CIs). CPU-to-NPU speedup is calculated for each matched seed before aggregation. Statistical comparisons employ paired or unpaired Wilcoxon tests as appropriate, Holm correction~\cite{Holm1979}, effect sizes, and Friedman ranks~\cite{Derrac2011Stats}.

Correctness is examined at component, generation, and repeated-run levels. Each common operator is compared with an independent source-consistent implementation on fixed tensors. Front numbers, feasibility masks, selected indices, neighbor orders, and truncation masks are required to match element by element. Continuous outputs include pairwise distances, crowding values, offspring tensors, and reference points. Their absolute and relative tolerances are fixed according to float32 before the output is inspected. Dedicated fixtures cover duplicate objective vectors, equal distances, zero objective ranges, empty constraint matrices, odd population sizes, non-finite values, and boundary repair.

Problem implementations are checked against their corresponding MATLAB definitions. The input set combines Latin-hypercube samples, lower and upper bounds, known Pareto points, and independently generated random decisions. The validation record contains output shapes, finite-value ratios, maximum absolute and relative errors, and feasibility mismatches. Permutation and assignment problems additionally require valid discrete structures and identical objectives on deterministic instances. This procedure detects errors that a short optimization run may conceal, such as an ignored decision variable, an unintended constant objective, or a constraint with the wrong sign.

Algorithm validation combines controlled intermediate states with indicator distributions. If two implementations consume the same parent indices and random-event tensors, the population, fitness, fronts, archive membership, and function-evaluation count are examined after every major calculation. Different language runtimes do not always expose identical random-number streams. In this case, repeated runs use common problem settings and compare HV, IGD, IGD+, IGDX, $1/\mathrm{PSP}$, feasible rate, and runtime distributions. A substantial difference is investigated through deterministic intermediate fixtures until the first divergent operation is identified. The source mapping and validation evidence for all 18 algorithms are reported in the supplementary material.

\subsection{Population Operators}

\begin{figure*}[t]
\centering
\includegraphics[width=0.99\textwidth]{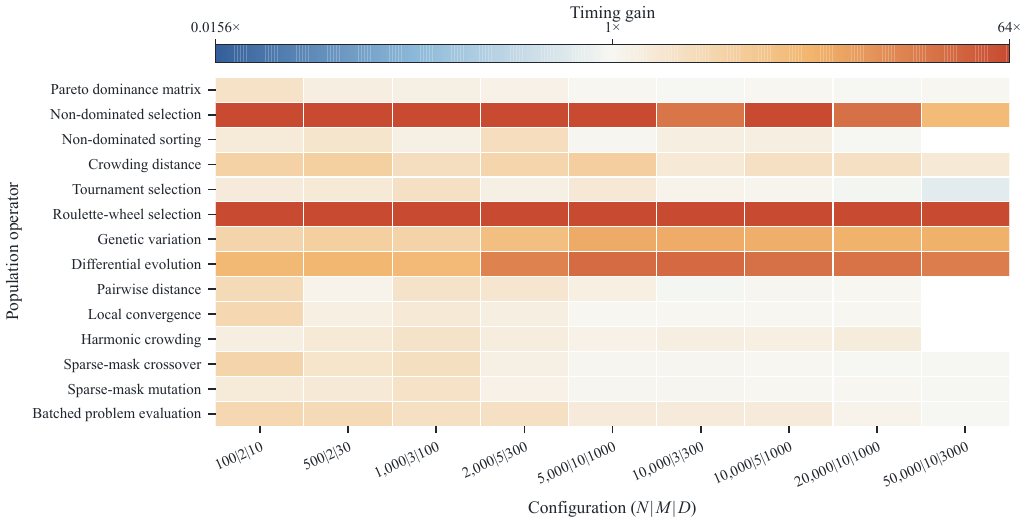}
\caption{Timing gain of the fully tensorized NPU operators relative to their source-consistent implementations. A value above $1\times$ favors tensorization, and the configuration labels follow $N|M|D$.}
\label{fig:operator_heatmap}
\end{figure*}

\Cref{fig:operator_heatmap} reports the gain obtained when the complete tensorized operator replaces its source-consistent NPU implementation. Each column represents one pressure configuration, and each row follows the standard name of a population operator. Across 122 matched configurations, the median gain is $1.58\times$, with an interquartile range of $1.14$--$6.94\times$. Non-dominated selection, roulette-wheel selection, differential evolution, and several batched evaluation settings show the largest improvements. Their calculations replace repeated scalar control with comparisons or reductions over complete tensors.

Some cells remain near parity because short operations contain too little arithmetic to offset launch and synchronization costs. The device comparison in the supplementary material shows a similar scale effect. Dense pairwise relations, genetic variation, differential evolution, and population evaluation increasingly favor the NPU as $N$ grows, while selection operators with little work per element cross over later. These results support the runtime model in \Cref{eq:runtime} and show why tensorization and device acceleration need to be evaluated separately.

\subsection{Performance at Different Scales}

The standard experiment contains 20,100 paired float32 runs from 1,005 algorithm--problem clusters at $N=100$ and $FE_{\max}=10{,}000$. The median CPU-to-NPU ratio is $0.980\times$, and the interquartile range is $0.923$--$1.026\times$. The clustered-bootstrap 95\% CI for the median is $0.976$--$0.983\times$. Hence, the NPU and eight-thread CPU remain close at this small workload, and the CPU has a slight median advantage. This outcome is consistent with the fixed launch and synchronization cost in \Cref{eq:runtime}.

The scale experiment contains 373 successful matched pairs from 77 algorithm--problem--configuration clusters. Population size ranges from $1{,}000$ to $20{,}000$, and $FE_{\max}$ ranges from $1\times10^5$ to $2\times10^6$. The median CPU-to-NPU speedup reaches $5.76\times$, with an interquartile range of $1.72$--$19.14\times$. Resampling complete clusters gives a 95\% CI of $3.26$--$9.09\times$. The population-wise results in the supplementary material show that every MOP category moves above parity as $N$ increases. Larger populations increase the time spent on pairwise relations and batched evaluation, which distributes fixed device costs over more useful arithmetic.

The dispersion remains wide at the three largest population sizes. At $N=20{,}000$, the aggregate first and third quartiles are $1.01\times$ and $32.26\times$, respectively. Different calculation structures explain much of this range. The number of successful pairs decreases from 85--86 at the three smaller population sizes to 71 and 46 at $N=10{,}000$ and $20{,}000$. Therefore, capacity outcomes change the composition of the aggregate groups. The supplementary analysis retains the algorithm, MOP, configuration, and seed associated with each matched pair.

\begin{figure*}[t]
\centering
\subfigure[NSGA-II on DTLZ2 for general MOEAs.]{\includegraphics[width=0.48\textwidth]{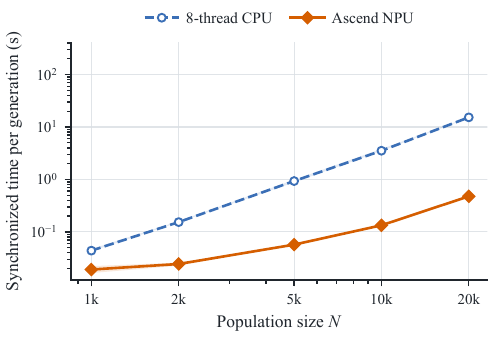}\label{fig:scale_general}}
\hfill
\subfigure[CMOEA-CD on CF1 for CMOEAs.]{\includegraphics[width=0.48\textwidth]{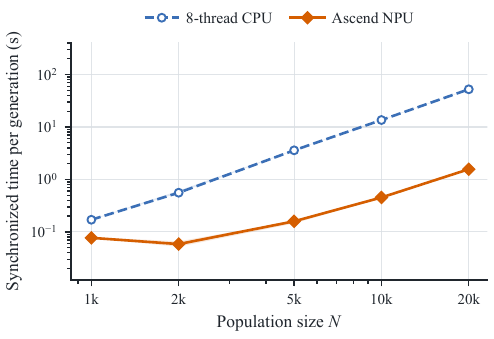}\label{fig:scale_constrained}}
\\[-0.5ex]
\subfigure[CoMMEA on MMF1 for MMOEAs.]{\includegraphics[width=0.48\textwidth]{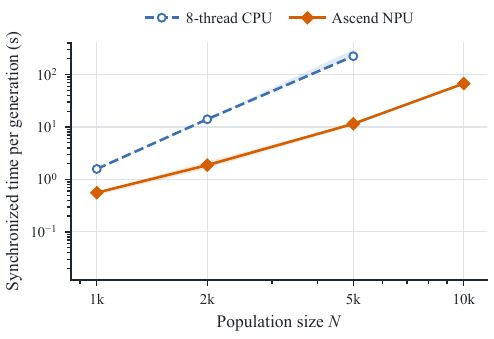}\label{fig:scale_multimodal}}
\hfill
\subfigure[LMEA on SMOP1 for sparse and large-scale MOEAs.]{\includegraphics[width=0.48\textwidth]{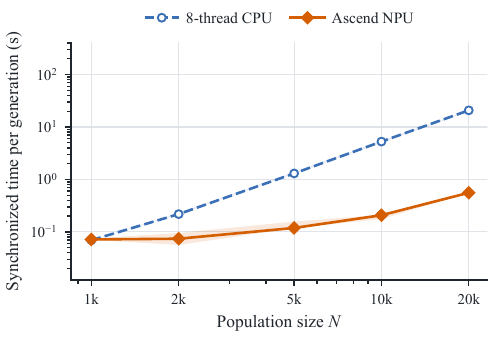}\label{fig:scale_sparse}}
\caption{Representative scaling of the synchronized time per generation for four MOEA categories. Lines and shaded regions show medians and interquartile ranges over five float32 seeds. Each value is the synchronized solve time divided by $(FE-N)/N$. A missing device point indicates that none of its five runs finishes successfully at that configuration.}
\label{fig:representative_scale}
\end{figure*}

\Cref{fig:representative_scale} presents the scaling process for one representative MOEA--MOP pair from each algorithm category. For each run, the synchronized solve time is divided by $(FE-N)/N$, which gives the number of completed population-sized updates after initialization. Each plotted device--configuration point reports the median over five float32 seeds, and the shaded region gives the interquartile range. The curves contain only observed successful runs, so a successful result from one device remains visible when the corresponding runs on the other device reach a resource limit. The CPU curves increase rapidly because pairwise relations and population calculations expand with $N$. The NPU curves grow more gradually after the initial device cost is covered.

The four panels reveal that the crossover scale depends on the calculation performed by each MOEA. At the largest successful matched population for each pair, NSGA-II~\cite{Deb2002NSGAII} on DTLZ2~\cite{Deb2002DTLZ}, CMOEA-CD~\cite{Liu2025CMOEACD} on CF1~\cite{Zhang2008CF}, CoMMEA~\cite{Li2023CoMMEA} on MMF1~\cite{Liang2019CECMMOP}, and LMEA~\cite{Zhang2018LMEA} on SMOP1~\cite{Tian2020SparseEA} reach median speedups of $31.38\times$, $33.56\times$, $21.16\times$, and $37.31\times$, respectively. The CoMMEA panel retains the successful NPU observations at $N=10{,}000$, although all five CPU runs at this configuration reach the 14-h time limit. At $N=20{,}000$, the CPU runs reach the time limit and the NPU runs terminate with OOM, so neither device contributes a point. LMEA remains close to parity at $N=1{,}000$, whereas the gap becomes clear at larger populations. Hence, the scale advantage occurs in every MOEA category represented by the experiment, but it does not begin at one universal population size.

\begin{figure}[t]
\centering
\includegraphics[width=0.47\textwidth]{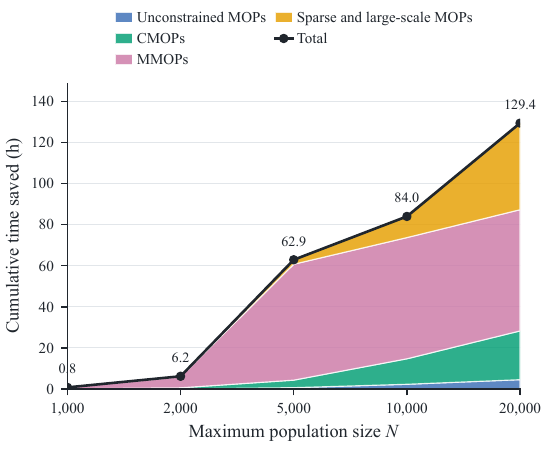}
\caption{Cumulative synchronized compute time saved by the NPU over successful matched scale runs. Each colored area gives the contribution of one MOP category, and the black line gives the total.}
\label{fig:time_saved}
\end{figure}

\Cref{fig:time_saved} expresses the same matched runs as cumulative compute time saved. For every successful pair, the synchronized NPU time is subtracted from its CPU counterpart, and the differences are accumulated by MOP category. The total reaches 129.4~h after configurations up to $N=20{,}000$ are included. MMOPs contribute 59.0~h, and sparse or large-scale MOPs contribute 42.1~h. The remaining 28.3~h comes from unconstrained MOPs and CMOPs. This decomposition reflects the duration of each workload together with its relative speedup.

\subsection{Effects of the Two Tensorization Levels}

The ablation separates common population operators from calculations located inside each algorithm. On the CPU, replacing the reference operators and then tensorizing the remaining calculations produce median gains of $1.037\times$ and $1.067\times$, respectively. Together, the two changes yield a median gain of $1.122\times$. Replacing only the common operators on the NPU is approximately neutral at $0.984\times$. Tensorizing the calculations inside each algorithm then raises the median gain to $1.363\times$, and the complete implementation is $1.295\times$ faster than the reference.

\begin{figure}[t]
\centering
\includegraphics[width=0.47\textwidth]{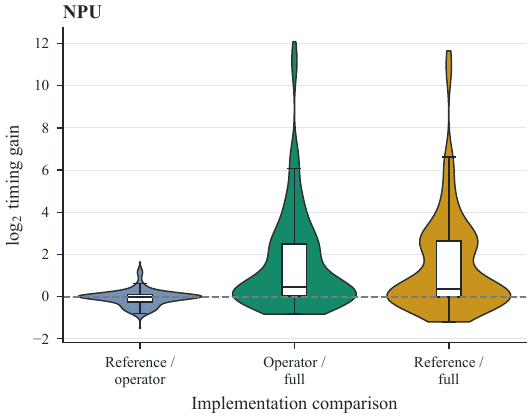}
\caption{NPU timing gains from common-operator replacement and tensorization inside each algorithm. Violin envelopes show the distributions, internal boxes show interquartile ranges, and zero on the $\log_2$ scale denotes parity.}
\label{fig:ablation}
\end{figure}

\Cref{fig:ablation} shows the complete NPU distributions behind these medians. The wide upper ranges occur because environmental selection and batched problem evaluation dominate some MOEA--MOP combinations. Tensorizing only a shared dominance or distance function cannot remove repeated indexing, scalar decisions, or individual evaluations that remain in the surrounding algorithm. The improvement becomes clearer when common operators and the calculations inside each algorithm are transformed together.

\begin{table}[t]
\centering
\footnotesize
\caption{Aggregate synchronized timing results for the full float32 implementation.}
\label{tab:timing}
\resizebox{0.47\textwidth}{!}{%
\begin{tabular}{lrrrr}
\toprule
Experiment & Pairs & Clusters & Median & 95\% CI \\
\midrule
Standard workload & 20,100 & 1,005 & $0.980\times$ & $0.976$--$0.983\times$ \\
Scale workload & 373 & 77 & $5.76\times$ & $3.26$--$9.09\times$ \\
Real-world subset & 2,920 & 146 & $0.965\times$ & $0.959$--$0.972\times$ \\
\bottomrule
\end{tabular}}
\end{table}

\subsection{Numerical Agreement and Practical Workloads}

Numerical agreement is evaluated independently of runtime. Among all paired standard runs, 19,479 pairs contain finite HV values, 19,478 contain finite IGD values, and 17,023 contain finite IGD+ values. The median and maximum relative CPU--NPU differences are zero at the stored precision for all three indicators. The supplementary identity plots show paired HV, IGD, and IGD+ values over their complete numerical ranges.

Terminal indicators can hide a transient difference, so complete population histories are examined. The supplementary material reports the absolute histories for NSGA-II~\cite{Deb2002NSGAII}, C-MOEA/D~\cite{Jain2014CMOEAD}, HREA~\cite{Li2023HREA}, and SparseEA~\cite{Tian2020SparseEA}. These examples cover IGD+, feasible rate, IGDX, and $1/\mathrm{PSP}$. The CPU and NPU medians overlap throughout the evaluation budget, and their interquartile regions remain consistent. Therefore, the terminal comparisons and complete histories support the discrete and continuous criteria in \Cref{eq:semantics} at both operator and algorithm levels.

The real-world subset contains 2,920 paired runs from 146 algorithm--problem clusters. Its median speedup is $0.965\times$, with a 95\% clustered-bootstrap CI of $0.959$--$0.972\times$. Most of these experiments use $N=100$, and several engineering or combinatorial objectives contain irregular gather operations with limited arithmetic intensity. Thus, their primary contribution is problem coverage and numerical comparability. The scale experiments provide a more direct measurement of NPU throughput under dense workloads.

Memory behavior is reported with the terminal state. Among the observed scale tasks included in the capacity analysis, 2,354 terminate successfully, 127 reach the 14-h limit, and 120 terminate with OOM. OOM and timeout are retained as outcomes rather than converted into zero time or missing quality. The supplementary material reports host and NPU memory growth together with completed-run counts. These records identify the practical boundary of full pairwise relations, whose storage grows quadratically with $N$.

\subsection{Discussion}\label{sec:discussion}

The experiments lead to three practical observations. Firstly, tensorization and accelerator execution are separate effects. Batched operators improve the CPU implementation as well, while an NPU provides an additional advantage only after dense computation becomes large enough to cover launch and synchronization costs. Therefore, a small standard workload remains close to parity. Secondly, complete algorithm structure matters more than the nominal problem category. Two algorithms with the same $N$ and $FE_{\max}$ can have different speedups because one spends more time on pairwise tensors and the other contains a longer ordered recurrence. Thirdly, numerical equivalence requires explicit control of ties, random events, precision, and write order. Indicator similarity alone cannot reveal a changed environmental selection rule.

The Ascend execution strategy provides a useful compromise for algorithms with mixed computational structure. Dense dominance, distance, variation, and objective calculations remain on the NPU. Compact precomputed data support the short recurrences that require strict ordering. This arrangement keeps NPU tasks in float32 and limits transfers to indices, masks, or sorted relations. It avoids repeatedly launching conditional kernels with little arithmetic.

However, full pairwise relations have $O(N^2M)$ computational and memory requirements, which restrict the largest feasible population. Some combinatorial objectives contain irregular memory access and cannot fully utilize matrix units. Float32 comparisons can be sensitive when objective or constraint values differ near machine precision, although the current validation detects changes in the resulting discrete state. The present platform targets one host with multiple Ascend devices. Distributed multi-node execution and compiler-level operator fusion offer further opportunities for scalability.

The current evidence suggests a practical selection rule. CPU execution is appropriate for small populations, short runs, and highly ordered algorithms. Ascend NPU execution becomes preferable when a generation contains large pairwise relations, high-dimensional variation, or batched problem evaluation. The synchronized crossover should be measured for a representative workload because device labels alone do not predict performance.

\section{Conclusion}\label{sec:conclusion}

This study presents \ascendmoea{}, which is a scalable platform for tensorized multi-objective evolutionary optimization on Ascend NPUs. A unified population representation connects algorithms, problems, operators, indicators, monitors, and experiments. Common calculations and the calculations defined inside each algorithm are expressed as population tensors, while updates with causal dependencies retain their prescribed sequence. The device policy provides explicit precision, seeding, synchronization, and timing behavior. The platform contains 18 MOEAs that span general MOEAs, CMOEAs, MMOEAs, and sparse or large-scale MOEAs. Its 155 MOPs cover unconstrained MOPs, CMOPs, MMOPs, sparse MOPs, large-scale MOPs, real-world MOPs, and combinatorial MOPs. Experiments show that tensorized operators obtain a median NPU gain of $1.58\times$. Complete algorithms remain near parity for a small standard workload and achieve a median speedup of $5.76\times$ at larger scales. Paired terminal indicators and generation histories retain the recorded numerical behavior. These results indicate that Ascend acceleration is most effective when dense population calculations dominate the evolutionary process. 

Future research can reduce the memory demand of dense pairwise relations through sparse or implicit representations that preserve the resulting dominance and selection decisions. Extending the runtime across multiple Ascend hosts is another important direction, and this execution model requires reproducible random states, coordinated synchronization, and efficient exchange of population tensors. Compiler fusion and persistent kernels can reduce launch costs for short selection and repair operations, whose arithmetic remains insufficient to occupy the NPU at small $N$. For expensive engineering MOPs, batched surrogate models and asynchronous evaluation can reduce costly objective evaluations while exposing additional parallel calculations. These extensions can be examined with the operator, scale, numerical agreement, and memory criteria established in this study.

%\section*{Acknowledgment}
%Funding information is omitted for double-anonymous review.

\small
\bibliographystyle{IEEEtran}
\bibliography{ref}

\end{document}